# Piecewise linear regressions for approximating distance metrics

Josiah Putman, Lisa Oh, Luyang Zhao, Evan Honnold, Galen Brown, Weifu Wang, Devin Balkcom

*Abstract*— This paper presents a data structure that summarizes distances between configurations across a robot configuration space, using a binary space partition whose cells contain parameters used for a locally linear approximation of the distance function. Querying the data structure is extremely fast, particularly when compared to graph search required for querying Probabilistic Roadmaps, and memory requirements are promising. The paper explores the use of the data structure constructed for a single robot to provide a heuristic for challenging multi-robot motion planning problems. Potential applications also include the use of remote computation to analyze the space of robot motions, which then might be transmitted on-demand to robots with fewer computational resources.

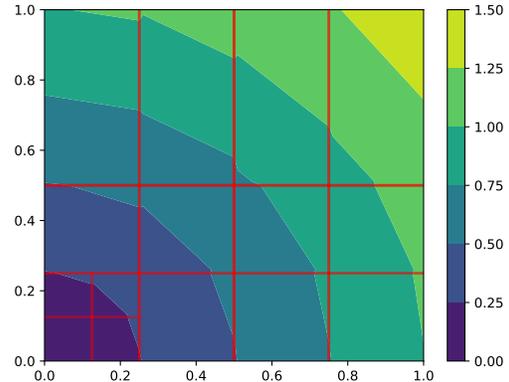

Fig. 1: A PLR of Euclidean distance to the origin.

## I. INTRODUCTION

As greater computational resources become available through large-scale clusters and cloud computing, the question arises of how to leverage those resources to allow fast and effective planning on a remote robot with far fewer resources. This paper takes one approach to the problem, finding approximate *compressed representations of optimal motion* that can then be stored, transmitted, and used within a tight computational budget. In particular, the paper discusses how to build and make use of a cell-based piecewise linear regression (PLR), where each cell contains a linear approximation of the value function.

We are particularly interested in motions that are optimal with respect to time cost, energy, precision, sensor coverage, or other objectives. In order to create a data structure that represents optimal motion, we must have: a) a method for discovering information about optimal motion, b) a method for storing that information, and c) a method for extracting optimal trajectories from the data structure.

Probabilistic Roadmap (PRM) frameworks provide all three methods: points are sampled and connected to analyze motion, and used to create a graph data structure from which paths may be extracted. The PRM* [13] algorithm converges to optimal paths in the limit, but convergence proofs require samples to be placed densely enough over the space to approximately cover any potential optimal path. This density can lead to a graph that is too large to store on disk or transmit over a network, and the computational cost of the A* search to find paths grows with the number of samples.

In the present paper, the value function is constructed primarily to summarize an existing roadmap data structure. As the computation of the initial roadmap is quite expensive, a promising direction of future work is the incremental construction of the PLR data structure without an existing graph.

The PLR data structure may represent the cost to a single goal, or may summarize an all-pairs distance function over the space. As an initial demonstration of the potential usefulness of the very high speed queries, we make use of the PLR as a heuristic for informed search using a traditional cell-based search method by Barraquand and Latombe [2]. While we do not claim that the informed search method we present is practically competitive with modern multi-robot planning methods, we believe that the high-speed distance function computation may serve as a useful component in future planning approaches.

### A. A simple example

Consider a toy problem that illustrates the challenges for PRM*, and the main insight that leads to approximation approaches. Let there be a point robot restricted to the box $[0, 1] \times [0, 1]$. For now, assume that the goal of the robot is always to reach the origin; we will relax this assumption shortly. Further assume that the true cost of reaching the origin is Euclidean: $d(x, y) = \sqrt{x^2 + y^2}$. To approximate this cost well, PRM* needs to place very many samples: for each possible starting configuration, there must be samples sufficiently close to an optimal trajectory such that the local planner can connect samples without deviating too far from the optimal.

For this toy problem, the analytical distance function may be computed quickly with high accuracy, and the formula requires little memory to store. But let us imagine that $d$ is a black box that may only be queried at particular points, and will later be unavailable to us. Let $p_1 = (0, 0, 0)$, $p_2 = (1, 0, 1)$, and $p_3 = (1, 1, \sqrt{2})$, where the first two elements of each vector give the $x$ and $y$ location of each point, and the third element gives the value of the distance function computed at that point. These three points describe a plane in $R^3$. Given a new starting configuration $(x, y)$, we may intersect the vertical line through $(x, y)$ with that plane to approximate the distance function.

Of course, the further we get from the known points, the greater we expect the error to be. To mitigate this issue, we divide the space into regions, with a different linear approximation in each region. We separate the problem into *construction* and *query* phases. To *construct* the desired data structure, we will use a binary space partition (BSP) to segment the space into cells. Within each cell, we construct a linear approximation of the distance function, by sampling the true distance function at a few points within or near the cell and computing parameters that are stored in the portion of the data structure corresponding to the cell. To *query* the approximate distance function at a point, identify the cell containing the point using the BSP and compute a dot product with the parameters associated with the cell.

## II. RELATED WORK

Perhaps the work closest to that proposed is on learning heuristics for robot motion planning in games by Rayner, Bowling, and Sturtevant [23], which attempts to remap a motion problem with obstacles into a new map for which the Euclidean distance represents a provably consistent, admissible heuristic for the original problem. Network embedding problems similarly try to find a mapping that expresses distance between vertices in a network; [10] provides a recent survey.

Work on LQR trees [26] places controllers over the state space, effectively reducing the memory requirements while also achieving safety of motion, providing a promising data structure to compute and transmit to robots. Bialkowski *et al.* have reduced the time cost of collision detection with RRT*, by building balls in free (Euclidean) space in which collision detection needs to be performed only once [7]. Deits *et al.* showed a numerical optimization approach to computing large convex regions, also in a Euclidean space [11]. Early work on neural network approximations of value functions derived from optimal control includes [20]. Recent work on distance metric approximation for RRTs using supervised learning [6] is also quite close in spirit to the proposed work, and was shown to be quite effective for finding a policy for a pendulum swing-up problem.

Like the present work, Probabilistic Roadmap (PRM) algorithms ([14]) algorithms divide motion planning into learning and query phases. When optimal paths are sought, as with the PRM* algorithm [13], roadmaps can become very dense. Marble et al. [18] introduced *spanners* [21, 24, 9, 27] into the robotics community to reduce the density of PRM* roadmaps (at some cost in path optimality) [16, 19, 15, 17]. In our work [28], a modified version of more recent streaming spanner algorithms [12, 4, 5, 25, 8] achieved similar path quality to the work by Marble in seconds rather than hours. Like spanner approaches, the present work attempts to find a summary of a graph data structure, but the summary is continuous within each cell.

## III. ALGORITHM

In this section, we define the algorithm by which the PLR is constructed over a metric space $Z$ using a black-box distance function to goal configuration $\mathbf{g} \in Z$, denoted by $d_g(\mathbf{x})$.

### A. Computing hyperplane coefficients

Given a set of $n$ sampled points $S \subset Z = \{\mathbf{x_1}, ..., \mathbf{x_n}\}$, we can write:

$$\mathbf{A} = \begin{bmatrix} \mathbf{x_1} & 1 \\ \mathbf{x_2} & 1 \\ \vdots & \vdots \\ \mathbf{x_n} & 1 \end{bmatrix} \mathbf{b} = \begin{bmatrix} d_g(\mathbf{x_1}) \\ d_g(\mathbf{x_2}) \\ \vdots \\ d_g(\mathbf{x_n}) \end{bmatrix}$$

Then solving the equation $\mathbf{Ac} = \mathbf{b}$ provides the coefficients $\mathbf{c}$ for the nearest fitting hyperplane to the points in $S$. An approximation of $d_g(\mathbf{x})$ can be computed using $L_g(\mathbf{x}) = \mathbf{c} \cdot \begin{bmatrix} 1 & \mathbf{x} \end{bmatrix}$.

### B. PLR construction algorithm

The following algorithm constructs a PLR over metric space $Z$ using analytic distance function $d_g(\mathbf{x})$ using the above coefficient computation, denoted as subroutine COMPUTECOEFFICIENTS.

The subroutine SHOULDSPLIT is the process by which the PLR decides when it should continue recursively splitting cells. This can be implemented in a variety of ways, the simplest of which is to split until some arbitrary maximum depth is reached. Alternatively, one can measure the error $\epsilon_c = |L(\mathbf{x_c}) - d_g(\mathbf{x_c})|$, where $\mathbf{x_c}$ is the center of the current cell and $L(\mathbf{x_c})$ is the PLR distance estimate at the given point $\mathbf{x_c}$, and split if $\epsilon_c$ is greater than some threshold $z$.

The SPLIT function splits a cell into two child cells along a given axis. For a cell split along dimension $n$ in and $N$ dimensional space, its children are split along the axis $(n+1) \mod N$.

**Data:** PLR $n_0$
$q \leftarrow [n_0]$;
**while** $|q| > 0$ **do**
    $n \leftarrow q.\text{pop}()$;
    COMPUTECOEFFICIENTS($n$);
    **if** SHOULDSPLIT($n$) **then**
        $l, r = \text{SPLIT}(n)$;
        $n.children.\text{append}(l)$;
        $n.children.\text{append}(r)$;
        $q.\text{append}(l)$;
        $q.\text{append}(r)$;
    **end**
**end**

**Algorithm 1:** PLR construction

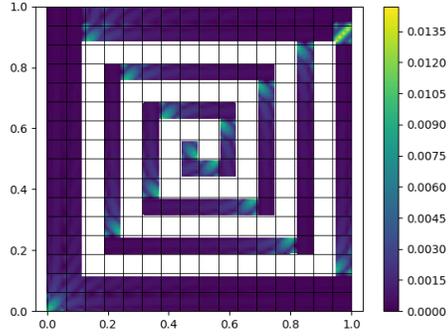

(a) $\epsilon_g(x)$ for a PLR constructed from a visibility graph.

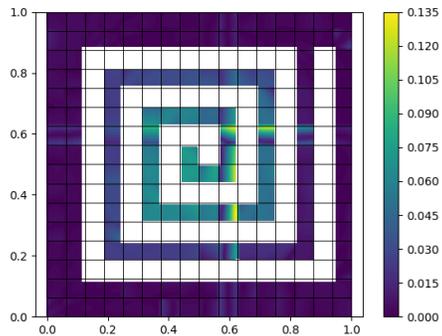

(b) $\epsilon_g(x)$ for a PLR constructed from a PRM*.

Fig. 2: PLR error functions for a point robot in a 2D maze

TABLE I: PRM-PLR Comparison

| Method | Memory (KB) | Max Error | Avg Error |
|---|---|---|---|
| PLR (VG) | 30.24 | 0.018 | 0.007 |
| PLR (PRM) | 30.24 | 0.134 | 0.016 |
| PRM* | 4412 | 1.472 | 0.024 |

## IV. PLR AS ESTIMATION OF NONLINEAR METRICS

Through experiments using PRM* and visibility graph as an analytic distance function in PLR construction, we demonstrate that PLR can provide an accurate summary of the distance function over a variety of spaces. Any metric space $Z$ for which a distance function $d_g(x)$ to some goal point $g$ is defined for all $x \in Z$ can have a PLR constructed to approximate its distance. The accuracy of the PLR depends on the optimality of $d_g(x)$.

### A. PLR over visibility graph

In the case of 2D spaces with polygonal obstacles, a visibility graph can be used to obtain a perfect analytic distance function to the goal. Because $d_g(x)$ is guaranteed to be optimal in this case, the PLR's accuracy can asymptotically approach the analytic distance function as its depth increases.

Figure 2a shows the error function $\epsilon_g(x) = |L_g(x) - V_g(x)|$, where $L_g(x)$ is a depth-9 PLR's distance estimate from $x$ to $g$ and $V_g(x)$ is the visibility graph's analytic distance, with obstacles indicated by the white region. The error at any given point in the cell is below 2%.

### B. PLR over PRM*

To construct a PLR over a higher dimensional space, the optimal analytic distance is often difficult or impossible to compute. A* search over a PRM* graph is a known method for retrieving a near-optimal distance to the goal in $n$-dimensional space with arbitrary obstacles. Given a sufficiently dense road map, a PLR can be constructed over the space using the length of an A* search result from the nearest point to the goal as its distance function. RPLRs constructed over PRM* graphs improve in accuracy with both the density of the roadmap and the depth of the PLR.

Figure 2b shows the error function $\epsilon_g(x) = |L_g(x) - V_g(x)|$, where $L_g(x)$ is a depth-9 PRM*-based PLR's distance estimate from $x$ to $g$ and $V_g(x)$ is the visibility graph's analytic distance. The error peaks at approximately 0.134 units, which is an order of magnitude smaller than the maximum error of the original PRM* it was built from. The PRM* has sparse regions at which there are very few samples, thus resulting in a path that is significantly suboptimal. As a linear approximation that is locally continuous in each cell, the PLR does not have these small areas of high error due to sparseness. The PLR also costs significantly less memory, taking only 30 kilobytes versus the 4.4 megabytes used by the PRM* (for about 10k samples).

## V. PLR AS A HEURISTIC

In this section, we demonstrate the PLR's effectiveness as a heuristic for known motion planning algorithms.

We first attempted to use PLR as heuristic with a modification of A* in continuous space, called Sampling-based A* algorithm (SBA*) [22]. In order to make progress in a continuous space, the authors introduced a local density function to push samples away from each other. However, though the algorithm follows a similar strategy as discrete A*, the introduction of different heuristics did not improve the performance of the algorithm in our experiments.

We then turned our attention to the algorithm introduced by Barraquand and Latombe [3] (which we refer to as BL planner). The algorithm uses a grid decomposition to push exploration of away from the already explored area. A heuristic can be used to direct the search towards the optimal path by using A* search to bias the order of exploration.

### A. PLR heuristic for single-robot motion planning

We compared the BL algorithm with and without a heuristic for a single robot in a environment with a small door connecting two large open regions. PLR was constructed in the environment using a visibility graph. The comparison result is shown in Table II. When using the PLR heuristic, the algorithm is much more efficient in finding a solution in the space and returns a near-optimal solution.

In the same environment, we also used RRT* to find a path of similar quality. RRT* used about the same time using fewer vertices, due to the fact that RRT* must rewire the graph as the algorithm progresses. This comparison demonstrates that PLR functions as an effective heuristic to the BL planner, providing comparable performance to RRT* in the case of single-robot motion planning.

### B. PLR heuristic for multi-robot motion planning

In this section we explore the effectiveness of the PLR heuristics for multi-robot motion planning. In the following experiments, the robots will have competing sections along respective optimal paths. We use the PLR constructed for individual robot as the heuristic for the BL algorithm. The motion cost for the multi-robot system is computed as the summation of the motion cost for each individual robot.

We conducted experiments first in an environment with four rooms connected by a common region in the center of the space. The entrance to the common region is tight for the disc robots. The comparison is shown in Table II. We compared the BL algorithm using PLR heuristic for single robot against RRT* with two robots, three robots, and four robots. In all cases, the BL algorithm equipped with the PLR heuristic produced paths of similar quality to RRT* in significantly less time.

We then conducted experiments in an environment where the free space is the union of two long hall ways. The disc robots needs to pass each other to get to different ends of hall way. The comparison is shown in Table II. When there are only two robots, RRT* is more efficient than BL with PLR heuristics. We hypothesize that this is because the space is very tight, and RRT* does not need to explore far to find a path. As the number of robots increases, the RRT* needs to explore more space to find a path, while BL planner remains efficient with the guidance of the PLR heuristic. Figures 3a and 3b shows a path found by the BL planner for three disc robots. The solid-color robots show the location of the robots at the end of the respective sequence.

Finally, we conducted experiments for two rectangle robots to pass each other in a narrow hallway. The middle of the hallway is more open, and the two rectangles must rotate around each other there to pass. The comparison is shown in Table II, and a resulting path found by BL planner is shown in Figure 3c. The BL planner outperformed RRT*, which failed to find a solution within a 30-minute time constraint.

Figure 4 shows the samples placed by the BL algorithm in these environments. The samples clearly show the BL planner utilizes the heuristic to find good quality paths near optimal solution.

In all multi-robot systems above, the RPLRs are generated using visibility graphs for a point even though the robots are not point robots, therefore providing a heuristic that is only correct in the global sense. Still, the correct global trend is sufficient to guide multi-robot planning problems to be solved more efficiently.

TABLE II: Comparison between BL planners with or without heuristic against RRT* in different environments.

|  | BL without heuristic | BL with PLR | RRT* |
|---|---|---|---|
| Single door | 1.46s (24864 samples) | 0.08s (3800 samples) | 0.08s (2200 samples) |
| Four rooms with 2 robots | N/A | 1.08s (991 samples) | 1.63s (1800 samples) |
| Four rooms with 3 robots | N/A | 26.23s (4514 samples) | 41.02s (17632 samples) |
| Four rooms with 4 robots | N/A | 83.26s (9407 samples) | 2762.38s (167216 samples) |
| Cross with 2 robots | N/A | 6.75s (936 samples) | 0.52s (402 samples) |
| Cross with 3 robots | N/A | 29.96s (2868 samples) | 94.71s (382 samples) |
| Tangle with 2 robots | N/A | 745.99s (19959 samples) | No solution after 7200 seconds |

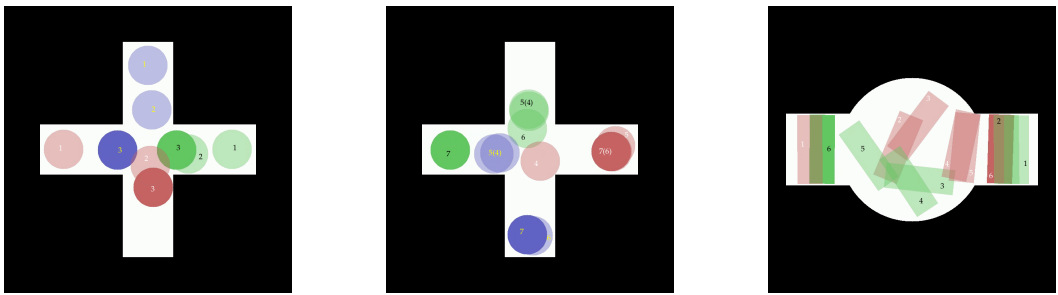

(a) The first half of the paths for the three discs.
(b) The second half of the paths for the three discs.
(c) The paths for the two rectangle robots.

Fig. 3: Paths for planner results.

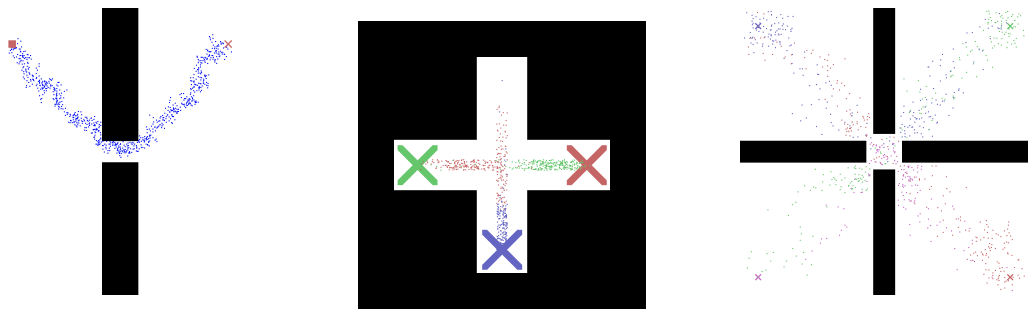

(a) The samples placed by BL planner (using PLR heuristic) in the single door environment.
(b) The samples placed for each of the three discs in the cross environment.
(c) The samples placed for each of the four robots in the four rooms environment.

Fig. 4: Samples placed by the BL planner using PLR as heuristic in different environments planning for different number of robots.

## VI. APPROXIMATION QUALITY ANALYSIS

In this section, we evaluate the approximation quality of the PLR's distance estimate. We prove that within cells that are not too close to the obstacles, the PLR approximation has bounded error.

First, we need to define when is a cell not too close to the obstacle. In particular, we adopt some of the terminology used in [1]. Given sets $X$ and $Y$ of a metric space, under a steering method $S$, we say that $X$ is path **subconvex** to $Y$ under $S$ if between any two points in $X$, the corresponding path in $S$ is contained within $Y$. Then, we have

*Theorem 1 (Theorem 1 from [1]):* Given an optimal steering method $S$, a corresponding metric $d$ over a metric space $Z$, a point $x \in Z$, and a positive constant $r$, the closed metric ball $B_{r/2}^d[x]$ centered at $x$ with radius $r/2$ is path subconvex to $B_r^d[x]$ under $S$.

In this work, cells are metric cubes rather than metric balls. One can extend the Theorem 1 to metric cubes rather than balls, by considering the parameter dimension independently. At the same time, as a metric cube with edge length $\epsilon$ is fully contained within a metric ball of radius $\epsilon/\sqrt{2}$, we can directly apply the theorem.

Let us denote the value function under the optimal steering method $S$ as $V(\cdot)$, and let $L(\cdot)$ be the approximated value function generated by PLR. Let $\{p_1, p_2, \ldots, p_n\}$ be the base points in a cell that are used to generate function $L(\cdot)$. Without loss of generality, let us assume that $V(\cdot)$ is Lipschitz continuous with constant $\kappa$. We can derive the following Lemma,

*Lemma 1:* Given a $n$ dimension PLR cell $C$ of edge length $\epsilon$ that is path subconvex under $S$, if the value function $V(\cdot)$ under $S$ is Lipschitz continuous with constant $\kappa$, then $|V(p) - V(q)|$ is upper bounded by $\kappa\epsilon\sqrt{n}$ for all $p, q \in C$.

*Proof:* Because $C$ is path subconvex, the value function $V$ satisfies triangle inequality. Then, for any given $p, q \in C$, we have,

$$V(p) \leq V(q) + V(p, q) \quad . \tag{1}$$

Along with the fact that $V$ is Lipschitz continuous, let $d(\cdot, \cdot)$ be the distance function in the parameter space, we have

$$|V(p) - V(q)| \leq V(p, q) \leq \kappa d(p, q) \quad . \tag{2}$$

For any pairs of points in a cell $C$ of edge length $\epsilon$ in dimension $n$, the maximum distance between them is $\sqrt{n} \cdot \epsilon$. Therefore, the value function difference between any two points in the cell is upper bounded by $\kappa\epsilon\sqrt{n}$. ∎

Then, given the set of base points $P = \{p_1, p_2, \ldots, p_n\}$ and the function $L$ generated using the base points, we can have the following theorem.

*Theorem 2:* Given a $n$-dimension PLR cell $C$ of edge length $\epsilon$ that is path subconvex under $S$, let $L(\cdot)$ be the PLR approximated value function generated from base points $P = \{p_1, p_2, \ldots, p_n\}$, if the value function $V(\cdot)$ under $S$ is Lipschitz continuous with constant $\kappa$, then $|V(p) - L(p)| \leq \frac{5}{2}\kappa\epsilon\sqrt{n}$.

*Proof:* Because $L$ is computed using the value functions at base points, the maximum slope on function $L$ is also $\kappa$. Because $C$ is path subconvex under $S$, we have triangle inequality for any given point $q \in C$,

$$\max_{p_i \in P} V(p_i) - \kappa\epsilon\sqrt{n} \leq V(q) \leq \min_{p_j \in P} V(p_j) + \kappa\epsilon\sqrt{n} \quad . \tag{3}$$

Because $L(\cdot)$ is a linear approximation, $L$ is monotonic. There are three possible cases for a point $q \in C$ and all $i$: 1) $L(q) < \min L(p_i)$; 2) $\min L(p_i) \leq L(q) \leq \max L(p_i)$; 3) $\max L(p_i) < L(q)$.

If $\min L(p_i) \leq L(q) \leq \max L(p_i)$, we have

$$|V(q) - L(q)| \leq |L(p_i) - V(p_i)| + \kappa\epsilon\sqrt{n}, \quad \forall i \tag{4}$$

Because $L$ is computed from $V$ at base points using least square approximation, we have $\max L(p_i) \leq \max V(p_i)$ and $\min L(p_i) \geq \min V(p_i)$ for all $i$. Then, the maximum difference between $V(\cdot)$ and $L(\cdot)$ at base points are upper bounded by $\kappa\epsilon\sqrt{n}/2$. Then, $|V(q) - L(q)| \leq \frac{3}{2}\kappa\epsilon\sqrt{n}$.

If $L(q) < \min L(p_i)$, we would also have $L(q) \geq \max L(p_i) - \kappa \max d(q, p_i)$ from triangle inequality. Then

$$|V(q) - L(q)| \leq |V(p_i) - L(p_i)| + \kappa\epsilon\sqrt{n} + \kappa \max d(q, p_i) \tag{5}$$

$$\leq \frac{1}{2}\kappa\epsilon\sqrt{n} + \kappa\epsilon\sqrt{n} + \kappa\epsilon\sqrt{n} \tag{6}$$

$$\leq \frac{5}{2}\kappa\epsilon\sqrt{n}. \tag{7}$$

If $\max L(p_i) < L(q)$, we can similarly derive $|V(q) - L(q)| \leq \frac{5}{2}\kappa\epsilon\sqrt{n}$. ∎

## VII. LIMITATIONS AND FUTURE WORK

The primary contribution of this work is conceptual: it suggests that the density of sample points produced by PRM*, RRT*, and related sampling algorithms may be avoided using simple direct approximations of value functions, reducing memory costs and increasing query speed dramatically. The approach is quite transparent relative to summarization based on neural networks, and may provide a good basis for comparison to learning-based approaches to motion planning.

However, there is much to be done before the PLR data structure is a practical component of a motion planning system. The most significant limitation is perhaps the cost of the construction phase. In the present paper, the issue was mitigated by computing the PLR for a single robot, and then using that PLR as a heuristic for a multi-robot motion planning problem. We are presently attempting to build PLR structures incrementally, connecting a few additional samples to the existing PLR to construct each cell; preliminary work is quite promising.

If incremental construction that avoids intermediate construction of a PRM* is computationally efficient, we expect to be able to increase the number of dimensions for which the approach is effective. However, we expect that the cell-based approach to become limiting for higher dimensions. We intend to explore different ways in which value function approximations may be built without relying on cells, perhaps with approximation methods that are smoother than the linear approach used.